\documentclass[11pt]{article}

\usepackage[margin=1in]{geometry}
\usepackage[T1]{fontenc}
\usepackage[utf8]{inputenc}
\usepackage{lmodern}
\usepackage{microtype}
\usepackage{amsmath,amssymb}
\usepackage{booktabs}
\usepackage{multirow}
\usepackage{graphicx}
\usepackage[numbers,sort&compress]{natbib}
\usepackage[colorlinks=true,linkcolor=blue,citecolor=blue,urlcolor=blue]{hyperref}
\usepackage{xcolor}
\hypersetup{
  pdftitle={Space-Creating versus Dead Possession: An Off-Ball Possession-Quality Index for Broadcast Football},
  pdfauthor={Seongjin Choi},
  pdfsubject={Possession quality, expected threat, space creation, spatial football analytics},
  pdfkeywords={football analytics, possession value, expected threat, pitch control, space creation}
}

\newcommand{\pp}{\,pp}   
\newcommand{\junk}{\textsc{junk-open}}
\newcommand{\eff}{\textsc{eff}}

\title{Space-Creating versus Dead Possession:\\An Off-Ball Possession-Quality Index for Broadcast Football}

\author{%
  Seongjin Choi\thanks{Independent researcher.
  ORCID: \url{https://orcid.org/0009-0001-3193-7424}.
  Code: \url{https://github.com/nowayfootball/junk-possession}.
  Contact: \texttt{nowayfootball@gmail.com}.}%
}
\date{}

\begin{document}
\maketitle

\begin{abstract}
Ball possession is the most-cited and most-misleading number in football:
sixty percent of the ball spent recycling in one's own half is not the same as
sixty percent spent pinning the opponent back. Existing event-based possession-value
frameworks --- expected threat, VAEP, on-ball value --- refine the count by
pricing \emph{on-ball} actions, but from event data alone they are blind to the
off-ball question a
sterile possession actually poses: did holding the ball \emph{create space},
or was the circulation genuinely dead? We answer this in two layers. First, an
event-side \emph{junk-possession index} decomposes possession into sequences,
prices each by its peak threat gain (credited even without a shot) under an expected-threat
grid, and --- after reconstructing the live scoreline to exclude
lead-protecting circulation --- flags low-threat sequences in tied-or-losing
game states. On the 2026 FIFA World Cup (103 matches, 206 team-matches) this
flag correlates negatively with points ($r=-0.37$) and expected-goal
difference ($r=-0.51$; the latter partly mechanically coupled, as the index
embeds a shot-xG term); team-matches above 55\% possession with below the
corpus-wide median efficiency average 0.62 goals and 1.00 points versus 2.24 and
2.24 for efficient dominators. Crucially, the flag is not a repackaging of
on-ball value: with team offensive VAEP and field tilt held fixed in the same
regression, the junk flag remains strongly negatively associated with points
($p<10^{-4}$, and under match-clustered errors) while VAEP itself is not
significant --- in this same-match (descriptive) regression the index adds
outcome-relevant information beyond this on-ball action-value model. Second, for
a flagged
window we resolve whether it was spatially dead or space-creating by projecting
broadcast video to pitch
coordinates and measuring a \emph{Space-Creation Index} (SCI): a net
pitch-control change capturing whether the possession seized dangerous space or
pushed the opponent's block back. Across 31 analysable flagged windows (of 35)
from nine purposively selected World Cup matches, 74\% are spatially classified
as non-space-creating,
19\% weak progression, and 6\% space-creating windows that the event flag alone
would score as failure --- including a possessing team that dominated 73\% of the
ball and went out on penalties (two windows, both spatially non-creating) set
against a space-creating window that produced no shot. The two layers together
separate ``space-creating but unconverted'' from ``sterile'' possession, a
distinction event-only on-ball possession value cannot make from its inputs
alone.
\end{abstract}

\section{Introduction}
\label{sec:intro}

Possession percentage is quoted after every match and understood by almost
no-one. It conflates three different things a team can do with the ball ---
where it holds the ball, whether that holding threatens the goal, and in what
game state it happens --- into a single number that is frequently uncorrelated
with, and sometimes inversely related to, winning. The analytics response has
been to \emph{value} what happens on the ball: expected threat
(xT)~\citep{singh2019xt} grids the pitch by scoring potential; VAEP~\citep{decroos2019actions}
and EPV~\citep{fernandez2021epv} estimate the change in scoring/conceding
probability around each action; on-ball value~\citep{statsbomb2021obv} and
pass-risk models~\citep{power2017passes,link2016dfl} price individual
touches. These are real advances, and they are all fundamentally
\emph{on-ball}: they evaluate the decision of the player in possession.

But the question a sterile possession poses is off-ball. When a team holds 65\%
of the ball, finishes level at 0--0, and loses the penalty shootout, we want to
know whether the possession was \emph{unlucky} --- it dragged the opponent's block out of shape
and manufactured dangerous space that went unconverted --- or \emph{dead} ---
the block never moved and the ball simply circulated. On-ball action value
cannot tell these apart, because the distinguishing evidence is the position of
the twenty-one players \emph{without} the ball: whether the defensive block was
compressed and displaced. That evidence is spatial, and for the vast majority
of matches --- outside the handful of leagues with commercial tracking --- the
only available source of it is the broadcast video.

This paper contributes an off-ball possession-quality framework in two layers:

\begin{enumerate}
\item \textbf{An event-side junk-possession index} (Section~\ref{sec:index})
that flags low-threat possession in decision-relevant game states, and a
validation battery (Section~\ref{sec:validation}) showing it is associated with
outcomes, holds up in held-out splits, and --- our central claim --- is
\emph{not reducible} to on-ball possession value: controlling for VAEP does not
remove its predictive content.
\item \textbf{A spatial Space-Creation Index} (Section~\ref{sec:sci}) computed
from broadcast video via a game-state-reconstruction (GSR) pipeline, which
adjudicates whether an event-flagged low-threat possession was spatially dead or
space-creating --- and a multi-match spatial study of 31 flagged windows from
nine World Cup matches (Section~\ref{sec:results}).
\end{enumerate}

The event layer scans whole matches cheaply and plants time-stamped flags; the
spatial layer answers ``why'' on the short broadcast windows the flags point
to. Neither layer alone suffices: event data cannot see the block, and
processing 90 minutes of broadcast video per match to pitch coordinates is
neither necessary nor, on commodity hardware, cheap. Our position is that the
macro/micro split is the practical way to bring off-ball spatial evidence to
possession analysis at the scale of a whole tournament.

\section{Related Work}
\label{sec:related}

\paragraph{Possession value.} Expected threat~\citep{singh2019xt} assigns each
pitch location a scoring potential and values ball progression between
locations. VAEP~\citep{decroos2019actions} and its EPV
predecessors~\citep{fernandez2021epv} learn, from event or tracking data, the
change in scoring and conceding probability that each action induces, thereby
valuing defensive actions, failed actions, and risk. On-ball
value~\citep{statsbomb2021obv} is a commercial analogue. All price the action
of the player on the ball; the event-only variants (xT, OBV, event-fed VAEP)
carry no off-ball player positions at all, while tracking-fed EPV can encode
defensive shape but needs full tracking, which broadcast does not provide. Our
index is complementary: Section~\ref{sec:vaep} shows empirically that, against an
event-fed VAEP, it captures outcome-relevant variance VAEP does not.

\paragraph{Sterile possession and territory.} ``Sterile domination'' and
``field tilt'' (here a proxy: a team's share of its own touches in the final
third, rather than the canonical share of both teams' final-third touches) are
established descriptive concepts in the analytics community. We show (Section~\ref{sec:partial}) that
field tilt explains expected goals but is not a significant predictor of points,
and that the junk flag carries points-relevant information beyond tilt --- possession quality and
territory are different axes.

\paragraph{Pitch control and space.} Pitch-control
models~\citep{spearman2017physics,spearman2018beyond} estimate, from player
positions, which team controls each point of the pitch; they require the
positions of all players. The Space-Creation Index is a differential
pitch-control statistic. Because broadcast video shows only a subset of
players, computing it from video invokes the off-screen-imputation problem
studied in our companion work; here we use the same CPU-only GSR pipeline and
flag imputation sensitivity as a limitation (Section~\ref{sec:limitations}).

\paragraph{Broadcast GSR.} Turning broadcast footage into pitch coordinates
combines camera calibration~\citep{gutierrez2024pnlcalib,falaleev2024keypoints},
multi-object tracking~\citep{aharon2022botsort}, and team assignment, evaluated
by GS-HOTA~\citep{somers2024soccernet}. Trajectory-imputation methods for the
players a camera or sensor misses~\citep{omidshafiei2022,everett2023inferring,
capellera2024transportmer} are learned and typically bidirectional; our
pipeline is training-free and online. We consume GSR output as a metric
substrate rather than advancing GSR itself.

\section{The Junk-Possession Index}
\label{sec:index}

\paragraph{Sequences and sequence value.} We segment each match into possession
sequences: maximal runs of consecutive events by one team; let $n_s$ be the
number of on-ball events in sequence $s$. Throughout, ``possession'' denotes a
team's share of on-ball events --- the standard event-data proxy --- so
``held 73\% of the ball'' is to be read as a 73\% on-ball-event share. Each
sequence $s$ is priced by
\begin{equation}
\mathrm{value}(s) = \underbrace{\max(0,\ \mathrm{xT}_{\max}(s) - \mathrm{xT}_{\mathrm{start}}(s))}_{\text{(i) peak threat gain (no shot needed)}}
+ 0.7\,\underbrace{\mathrm{xG}(s)}_{\text{(ii) realised shot}}
+ 0.10\,\underbrace{\mathbb{1}[\text{box touch}]}_{\text{(iii) any touch in the 18-yd box}},
\label{eq:value}
\end{equation}
where $\mathrm{xT}$ is a Karun-Singh-style expected-threat grid evaluated at
each touch location. Term (i) is the design core: a sequence is credited for
the \emph{peak} threat location it reached even if no shot resulted, so a
possession that worked the ball into a dangerous position but did not finish is
not scored as worthless. Term (ii) is the quality of the shot the sequence
produced, if any --- the \emph{maximum} shot xG among the shots in $s$; because
the event feed's shot-xG field is a constant placeholder, we substitute our own
gradient-boosted xG model (trained on our event corpus, distance/angle features
measured and body part/phase approximated; specification and fitted model
released in the repository) evaluated on shot location. Term (iii) rewards a
penalty-box touch --- a binary credit for any on-ball event inside the 18-yard
box during the sequence. Sequence values are divided
by the corpus-wide 90th percentile and clipped to $q(s) \in [0,1]$; a sequence
with $q < 0.15$ is a \emph{junk sequence} --- a possession of negligible threat.

\paragraph{Game-state normalisation.} Circulating in one's own half to protect
a lead is sound tactics, not junk. We reconstruct the live scoreline from goal
events and tag each sequence with the possessing team's goal difference at its
start. The headline metric \junk{} is the on-ball-event share of junk sequences
($q<0.15$) among a team's possessions in \emph{tied-or-losing} states
($\text{state} \le 0$) --- circulation with no excuse --- i.e.\
$\junk = 100\,\sum_{s\in T:\,\mathrm{state}(s)\le 0,\ q_s<0.15} n_s \big/
\sum_{s\in T:\,\mathrm{state}(s)\le 0} n_s$. Team-level aggregates
(Table~\ref{tab:metrics}) follow: raw possession, threat-weighted
\emph{effective} possession, efficiency
$\eff = \sum_{s\in T} n_s q_s / \sum_{s\in T} n_s$, \junk{}, field tilt, and a
sterile index.

\begin{table}[t]
\centering
\small
\begin{tabular}{lp{0.72\linewidth}}
\toprule
Metric & Definition \\
\midrule
Raw\% & Share of on-ball events \\
Effective\% & Threat-weighted event share, $100\,\dfrac{\sum_{s\in T} n_s q_s}{\sum_{s} n_s q_s}$ \\
Efficiency (\eff) & $\sum_{s\in T} n_s q_s / \sum_{s\in T} n_s \in [0,1]$ \\
\junk{} & Junk ($q<0.15$) event share among tied-or-losing possessions \\
Field tilt & Share of a team's on-ball events in its final third ($x\ge80$ on the 0--120 feed, coordinates oriented so the team attacks toward $x=120$); a field-tilt proxy \\
Sterile index & $(\text{Raw}-50)\times(1-\eff)$ \\
\bottomrule
\end{tabular}
\caption{Team-level possession-quality metrics.}
\label{tab:metrics}
\end{table}

\section{Validation}
\label{sec:validation}

We validate on 206 team-match rows drawn from 103 of the 104 matches of the 2026
FIFA World Cup group and knockout stages (one match is absent from our event
feed). Knockout ties decided on penalties are scored by their shootout result
(advance $=$ win), not as draws.

\subsection{Does junk possession relate to outcomes?}
\label{sec:corr}

Table~\ref{tab:corr} gives Pearson correlations of each possession metric with
match outcomes. \junk{} is negatively associated with expected goals
($-0.38$), goals ($-0.38$), points ($-0.37$), and xG difference ($-0.51$):
the more a team circulates without threat in a tied-or-losing state, the less
it scores and wins. (The xG and xG-difference columns are partly
mechanically coupled to the index, which embeds a shot-xG term ---
Section~\ref{sec:limitations}; goals and points are the cleaner columns.) Threat-weighting helps in this sample --- effective
possession tracks xG slightly better than raw possession ($0.56$ vs $0.50$) --- and efficiency is
the strongest single threat signal in this sample ($0.59$). Field tilt
correlates with xG ($0.58$) but weakly with points ($0.18$): camping in the final third without winning is precisely
what junk captures and tilt does not (Section~\ref{sec:partial}).

\begin{table}[t]
\centering
\small
\begin{tabular}{lcccc}
\toprule
Metric & Team xG & Goals & Points & xG diff \\
\midrule
Raw\%        & $+0.50$ & $+0.33$ & $+0.40$ & $+0.63$ \\
Effective\%  & $+0.56$ & $+0.37$ & $+0.42$ & $+0.70$ \\
Efficiency   & $+0.59$ & $+0.42$ & $+0.33$ & $+0.63$ \\
\junk{}      & $-0.38$ & $-0.38$ & $-0.37$ & $-0.51$ \\
Field tilt   & $+0.58$ & $+0.27$ & $+0.18$ & $+0.60$ \\
\bottomrule
\end{tabular}
\caption{Pearson correlation of possession metrics with match outcomes
(206 team-matches). \junk{} and efficiency measure possession \emph{quality};
field tilt measures territory.}
\label{tab:corr}
\end{table}

\begin{figure}[t]
\centering
\includegraphics[width=0.62\linewidth]{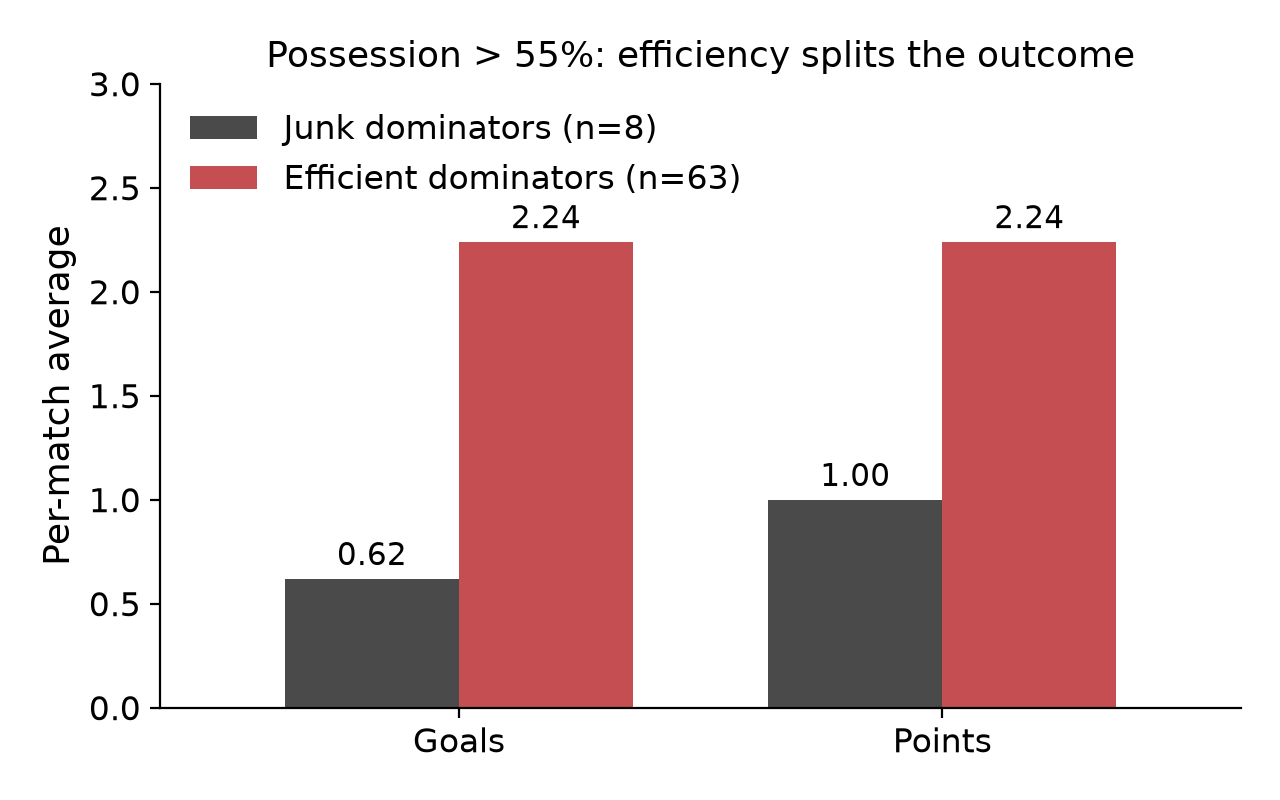}
\caption{Among the 71 team-match observations over 55\% possession, splitting by
corpus-wide median efficiency separates outcomes sharply: the 8 low-efficiency
``junk'' dominators average 0.62 goals and 1.00 points, versus 2.24 and 2.24 for
the 63 efficient dominators (eight observations, so descriptive only).}
\label{fig:quadrant}
\end{figure}

Splitting the 71 team-match observations above 55\% possession by efficiency
sharpens the point: the 8 that fell below the \emph{corpus-wide} median
efficiency (junk dominators) averaged 1.07 xG, 0.62 goals, and 1.00 points; the
63 above it (efficient dominators) averaged 2.05 xG, 2.24 goals, and 2.24
points. In this sample the eight averaged 28\% of the efficient group's goals
and draw-level points (Figure~\ref{fig:quadrant}; eight observations, so
descriptive only).

\subsection{Learned sequence weights}
A logistic model predicting whether a possession sequence ends in a shot, from
its peak threat gain, box touch, final-third share and touch count, reaches
5-fold AUC $0.907 \pm 0.004$ (mean $\pm$ cross-fold standard deviation; folds
are not match-grouped, so this is an upper-bound estimate). All threat features
load positively; the fitted standardised box-touch coefficient ($+0.60$) is
large relative to peak-threat gain ($+0.39$), which motivated raising the
box weight from the initial $0.03$ to $0.10$ in Eq.~\ref{eq:value} --- a
data-informed but not formally calibrated choice (we did not grid-search the
value).
Final-third share is deliberately \emph{excluded} from
Eq.~\ref{eq:value}: including it collapses the index toward field tilt, a
territory measure we deliberately keep distinct from possession quality
(Section~\ref{sec:partial}).

\subsection{Not a repackaging of territory}
\label{sec:partial}
Regressing points on raw possession, field tilt, and \junk{} jointly, \junk{}
remains significantly negatively associated with points ($\beta=-0.020$,
$p=0.017$) while field tilt is not ($p=0.14$); regressing xG difference, the
significance pattern reverses --- tilt is strongly significant ($p<10^{-4}$) and
\junk{} is absorbed ($p=0.77$). (Both are OLS; $p$-values here are iid ---
the \junk{}-on-points result also holds under match-clustered CR1 errors,
$p=0.014$.) In these two models, chance-creation (xG
difference) tracks territory while points track possession quality
--- consistent with the two being different axes, not a relabelling.

\subsection{Not reducible to on-ball possession value}
\label{sec:vaep}
The sharpest objection is that the junk index merely repackages on-ball
possession value (EPV, VAEP, OBV). We test this directly with our own VAEP
implementation~\citep{decroos2019actions} (learned on the same WhoScored event
schema; specification and code released in the repository). A weak comparator
could spuriously favour our index, so we note that this is an event-fed VAEP ---
the appropriate baseline for the event-only regime this paper targets. At
the surface, team offensive VAEP is \emph{correlated} with the possession-quality
metrics ($r=-0.49$ with \junk{}, $+0.58$ with efficiency, $+0.52$ with
effective possession) --- so the metrics are not orthogonal by construction;
VAEP is a genuine competitor. Yet when VAEP and field tilt are entered as
controls, the possession-quality metrics remain the significant predictors
while VAEP does not (Table~\ref{tab:vaep}): \junk{} is associated with points at
$p<10^{-4}$ with VAEP not significant ($p=0.34$), and efficiency with xG
difference at $p<10^{-4}$ with VAEP not significant ($p=0.46$). Despite a surface
correlation of $\approx0.5$, conditional on the possession-quality metric VAEP
is not a significant predictor of the outcome, so the junk index carries
outcome-relevant variance that this on-ball action-value model does not capture
on its own. This is the paper's central quantitative claim. (We report
conditional significance --- not a zero VAEP effect, nor a formal test that the
two coefficients differ; confidence intervals and a match-clustered
re-estimation are in the repository. And because \junk{} is constructed from
live score state, this same-match regression is descriptive
(Section~\ref{sec:limitations}); the cross-fitted splits below carry the
predictive weight.)

\begin{table}[t]
\centering
\small
\begin{tabular}{lccc}
\toprule
Outcome $\sim$ controls $+$ quality & term & $\beta$ & $p$ \\
\midrule
\multirow{2}{*}{Points $\sim$ VAEP $+$ tilt $+$ \junk{}}
  & VAEP  & $+0.29$ & $0.34$ \\
  & \junk{} & $-0.034$ & $<10^{-4}$ \\
\midrule
\multirow{2}{*}{xG diff $\sim$ VAEP $+$ tilt $+$ efficiency}
  & VAEP       & $+0.17$ & $0.46$ \\
  & efficiency & $+6.14$ & $<10^{-4}$ \\
\bottomrule
\end{tabular}
\caption{Possession-quality metrics survive controlling for on-ball VAEP; VAEP
does not survive controlling for them. (206 team-matches; full coefficient
table in the repository.)}
\label{tab:vaep}
\end{table}

\subsection{Out-of-sample}
In-sample correlations risk tautology (describing games already lost). Two
splits that do not use the target game's or period's own possession--outcome
relationship (up to the corpus-wide normalisation constants noted in
Section~\ref{sec:limitations}): (a) \emph{first-to-second
half} --- a team's first-half efficiency predicts its second-half xG
($r=+0.32$) and first-half \junk{} predicts lower second-half xG ($r=-0.24$),
a within-team comparison across the two halves (this does not adjust for latent
team strength --- the leave-one-match-out test below is a separate, complementary
probe, not a strength control); (b)
\emph{leave-one-match-out team trait} --- a team's mean efficiency in its
\emph{other} matches predicts this match's points ($r=+0.31$) and its mean
\junk{} predicts lower points ($r=-0.30$) and xG difference ($-0.32$).
Junk possession behaves like a team trait: its value in a team's \emph{other}
matches carries information about this match --- a cross-fitted association (using
both earlier and later matches), not merely a post-hoc description, though not a
strict time-respecting forecast.

\subsection{Dead versus set-up junk}
Extending the label across possession boundaries, a junk sequence is
\emph{redeemed} if the same team regains the ball within 25\,s and reaches
threat (a peak-xT gain $\ge 0.04$ or a box touch on that next possession) ---
a build-up, not dead circulation. Only 9\% of tied-or-losing junk (event-weighted,
the same denominator as \junk{}) is redeemed under this rule; the remaining 91\% is not. The
dead-only variant improves in-sample
prediction slightly (points $r$: $-0.37 \to -0.41$) but is essentially
identical out-of-sample --- an honest null: most flagged junk is not redeemed,
so the refinement adds interpretation, not power.

\section{The Spatial Layer: Space-Creation Index}
\label{sec:sci}

The event index sees only the ball; it flags \emph{that} a possession failed to
threaten, not \emph{why}. To adjudicate whether a flagged possession created
space, we project broadcast video to pitch coordinates and measure the change
in pitch control.

\paragraph{GSR pipeline.} We use a CPU-only broadcast pipeline of open
components: PnLCalib camera calibration~\citep{gutierrez2024pnlcalib} with a
temporal anchor bridge, BoT-SORT tracking~\citep{aharon2022botsort},
colour-based team assignment (per-match Lab chroma comparators, robust to
daylight), and shot-change gating to drop replays and close-ups. Off-screen
players are imputed with a training-free ghosting layer (companion work). The 31
analysed windows are those surviving kit- and clip-quality screening; we did not
additionally filter on imputation sensitivity, which we flag as a limitation
rather than a selection criterion.

\paragraph{Space-Creation Index.} For a flagged window with possessing team
$P$ and opponent $O$, let $\bar C^{\text{first}}$ and $\bar C^{\text{last}}$
denote a control share averaged over the first and last thirds of the window's
frames. With $A$ the attacking third (the third of the pitch nearest $O$'s goal) and
$Z$ the opponent's own attacking third (the mirror third, nearest $P$'s goal),
and pitch-control share aggregated over each zone,
\begin{equation}
\mathrm{SCI} = \Delta_{\mathrm{own}} + \Delta_{\mathrm{opp}},\qquad
\Delta_{\mathrm{own}} = \bar C^{\text{last}}_{P,A} - \bar C^{\text{first}}_{P,A},\qquad
\Delta_{\mathrm{opp}} = \bar C^{\text{first}}_{O,Z} - \bar C^{\text{last}}_{O,Z},
\end{equation}
so $\Delta_{\mathrm{opp}}>0$ means $O$'s presence in $Z$ \emph{receded} --- its
shape was pushed back (both terms in
percentage points, from the pitch-control model of
\citet{spearman2017physics}). SCI is a \emph{net} two-zone change: large positive
SCI $=$ the possession seized the attacking third and/or pushed the opponent's
block back (space-creating); SCI $\approx 0$ $=$ the
block held its shape; strongly negative SCI $=$ the possessing team \emph{lost}
attacking-third control while the opponent's block advanced --- circulation that
went nowhere or backwards. We use the prespecified operational bins from earlier
internal work: \emph{space creation} (SCI $\ge +12$), \emph{weak progression}
($+4 \le$ SCI $< +12$), and \emph{non-space-creating} (SCI $< +4$, spanning
static and regressive possession). All analysed windows already satisfy the
event-side low-threat definition ($q<0.15$); SCI does not revise that
event-side classification --- it determines whether each event-flagged low-threat
sequence was nevertheless spatially dead, weakly progressing, or space-creating.
Throughout, ``junk'' names the event-side candidate flag, not a final verdict.

\paragraph{Macro/micro roles.} The two layers divide labour
(Table~\ref{tab:layers}): the event index scans whole matches instantly and
plants time-stamped flags; the spatial layer processes the short broadcast
window each flag points to and answers why.

\begin{table}[t]
\centering
\small
\begin{tabular}{lll}
\toprule
Layer & Input & Role \\
\midrule
Event junk index & Full-match events (instant) & Time-stamp \emph{flags} junk \\
Spatial SCI      & 10--60\,s broadcast window   & \emph{Why} (was the block moved?) \\
\bottomrule
\end{tabular}
\caption{The macro (event) and micro (spatial) layers.}
\label{tab:layers}
\end{table}

\section{Multi-Match Spatial Results}
\label{sec:results}

The nine matches here are a \emph{purposive}, not random, sample: they were
chosen for a high sterile index (so that flagged junk windows exist), adequate
kit contrast, and late-round narrative interest, with a bias toward candidate
space-creation cases to stress-test the spatial layer. The proportions below
therefore describe this selected sample, not the tournament. Each event flag
carries the timestamp of its junk sequence; we map it to a broadcast window by
scoreboard-clock kickoff alignment and clip the surrounding 10--60\,s of
continuous wide-angle play (full clipping rule in the repository; extra time
excluded). We processed every such event-flagged junk window for nine 2026
World Cup matches: of 35 flagged windows, 31 survived the pipeline
(Table~\ref{tab:agg}). Across the nine analysed matches, \textbf{74\% of flagged junk is
spatially classified non-space-creating}, 19\% weak progression, and \textbf{6\%
space-creating} windows that the event flag alone would score as failure
(proportions from 31 windows; a small-sample estimate, not a tournament-wide
prevalence). The proportions shifted only modestly as this sample grew from six
matches (21 windows; 71/24/5) to nine (31; 74/19/6): most flagged junk in this
purposively selected sample is non-space-creating, and the spatial layer earns
its keep on the minority that receive a different spatial verdict.

\begin{table}[t]
\centering
\small
\begin{tabular}{lcccc}
\toprule
Match & Windows & Non-creat.\ & Weak & Space \\
\midrule
NED--MAR (R32) & 3 & 3 & 0 & 0 \\
ENG--GHA (grp) & 4 & 3 & 1 & 0 \\
NED--JPN (grp) & 4 & 3 & 1 & 0 \\
FRA--SWE (R32) & 4 & 4 & 0 & 0 \\
PAR--FRA (R16) & 2 & 1 & 1 & 0 \\
FRA--MAR (QF)  & 4 & 1 & 2 & 1 \\
GER--PAR (R32) & 3 & 2 & 0 & 1 \\
NOR--BRA (R16) & 4 & 4 & 0 & 0 \\
MEX--ENG (R16) & 3 & 2 & 1 & 0 \\
\midrule
Total          & 31 & 23 & 6 & 2 \\
\bottomrule
\end{tabular}
\caption{Spatial verdicts on event-flagged junk windows, nine World Cup
matches. Of 35 flagged windows, four were excluded (three for low kit contrast,
one for poor clip quality), leaving 31 analysed.}
\label{tab:agg}
\end{table}

\paragraph{Case: sterile domination that went out.} Germany held 73\% of the
ball against Paraguay in the Round of 32, were held level at 1--1, and lost the
penalty shootout to go out. Both flagged Germany windows are strongly negative
(SCI $-24.4$ and $-14.2$): the ball moved, but Germany \emph{lost}
attacking-third control while the Paraguayan block, far from being pushed back,
advanced. This is the textbook
the index is built to catch --- a scoreline (elimination) that possession
percentage alone would never predict from a dominant side, evident
off-ball. The outcome coding matters here: for our advancement-based outcome
definition, scoring the shootout elimination as a Germany loss preserves the
advancement information, whereas coding the match as a regulation draw --- a
legitimate alternative convention --- would attenuate this particular
association.

\begin{figure}[t]
\centering
\includegraphics[width=\linewidth]{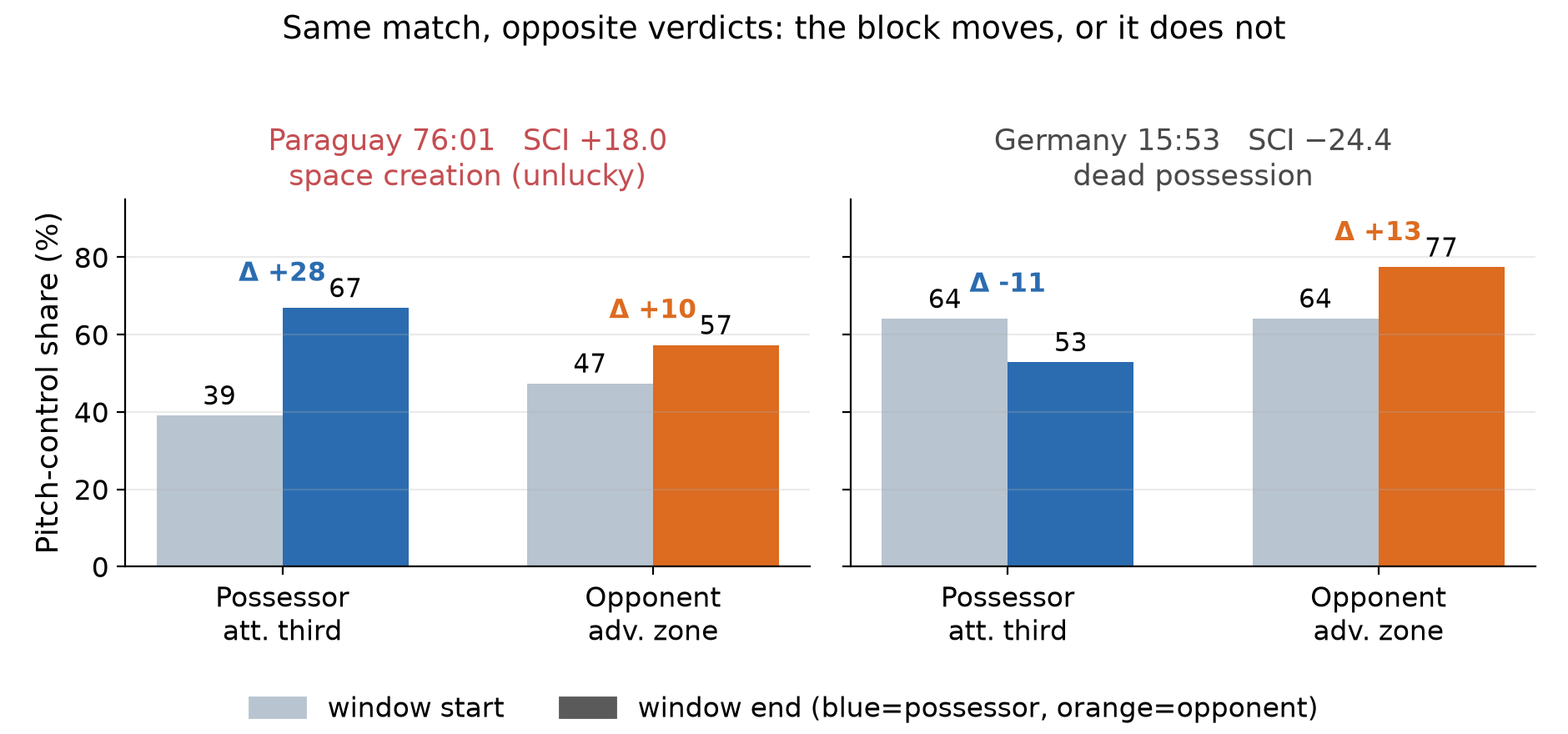}
\caption{Two flagged junk windows from the same match (Germany--Paraguay),
opposite verdicts. Pitch-control share averaged over the first versus last third
of the window. Left:
Paraguay creates space --- its control of the attacking third jumps 39$\to$67
(SCI $+18.0$). Right: Germany's possession is non-creating --- its
attacking-third control falls 64$\to$53 while Paraguay's block advances 64$\to$77
(SCI $-24.4$). $\mathrm{SCI} = \Delta_{\mathrm{own}} + \Delta_{\mathrm{opp}}$; the
event flag alone cannot tell the two apart.}
\label{fig:sci}
\end{figure}

\paragraph{Case: space-creating, not sterile.} In the same match, a Paraguay
window at 76:01 scores SCI $+18.0$ --- space-creating. The event index flagged
it as junk (no shot), but Paraguay seized the attacking third
($\Delta_{\mathrm{own}}=+28$\pp{}); the net SCI is positive even though the
opponent term worked against it --- Germany's presence in the mirror zone near
Paraguay's goal actually \emph{grew} ($\Delta_{\mathrm{opp}}=-10$\pp{}) --- so
the space was won in Germany's defensive third, not by pushing Germany's shape
back. Paraguay did not
convert it (whether through poor execution or chance, the spatial layer does not
adjudicate). This is the 6\% the two-layer design
exists for: the event flag reads low threat (failure) while the spatial verdict reads space creation, and the spatial
layer supplies the off-ball evidence the event flag cannot.
Figure~\ref{fig:sci} contrasts this window with a non-creating Germany window from the
same match: the possessing team's control of the attacking third surges
($+28$\pp{}) when it creates space, and falls ($-11$\pp{}) while the opponent's
block advances ($+13$\pp{}) when the possession is non-creating.

\paragraph{Case: junk flags on a winning side.} Norway beat Brazil despite a
high junk-flag rate; all four flagged Norway windows are spatially non-creating,
and Norway still won. High junk share and a win are not a contradiction: the
flags mark possessions that SCI likewise classifies as non-creating, and the index does
not claim those were the only possessions --- a team can waste much of the ball
and still convert elsewhere.

\paragraph{Event-side contrast.} The event layer already separates cases that
share a scoreline. South Africa lost 0--1 to Canada holding 57\% (efficiency
0.27, \junk{} 57\%, tied until the 91st minute): textbook junk, no excuse.
South Korea also held 57\% and lost 1--0 to Mexico, but at efficiency 0.41 ---
their possession was less sterile by this index --- whatever cost them the
match, this index does not locate it in circulation without threat. Same
possession, same result, opposite diagnosis.

\section{Limitations and Future Work}
\label{sec:limitations}

Sequences are segmented by possession change (a change in the on-ball team)
only; period boundaries, stoppages, and restarts are not special-cased, and
provider-coded opponent touches that do not transfer control can split a
possession --- a coarse segmentation that our aggregate metrics tolerate but that
a per-sequence study would need to refine. The event index is a ball-only
approximation of an off-ball question; the spatial layer supplies the off-ball
evidence only on short flagged windows, not whole matches --- a macro/micro hybrid, which we argue is the realistic optimum
for broadcast footage but is not full-match tracking. The xG term uses shot
location only (body part and phase approximated), and the sequence weights
(0.7, 0.10) and thresholds ($q<0.15$; SCI $+12/+4$) are round operational
values, partly justified by the learned weights (Section~\ref{sec:validation})
but not externally optimised. Two further construct caveats: sequence value
includes a shot-xG term (Eq.~\ref{eq:value}), so the xG-based associations
(Table~\ref{tab:corr}, the efficiency--xG-difference regression) are partly
mechanically coupled --- the \emph{goals} and \emph{points} outcomes are not
built into the index's formula, though they are not fully independent of it:
shot xG predicts goals, goals move the score state that gates \junk{}, and the
xG model itself is fit in-sample on this corpus's shot--goal labels (no
out-of-fold protocol), so goal information enters the index weakly through that
fit --- but they remain the cleaner endpoints; and the $q$-normalising
90th percentile, the $q<0.15$ threshold, and the learned box weight are all
corpus-wide quantities, so the out-of-sample splits are cross-fitted only up to
these global constants. On inference: \junk{} is built from
contemporaneous score state, so the in-sample outcome associations
(Section~\ref{sec:corr}) are descriptive and susceptible to score-state
feedback; the out-of-sample and leave-one-match-out splits
(Section~\ref{sec:validation}) are the genuinely predictive evidence. The 206
team-match rows form 103 paired matches (xG difference is antisymmetric within a
pair); we therefore re-estimated the headline regressions with match-clustered
(CR1) standard errors over the 103 clusters, and the central results are
unchanged --- \junk{} stays significant in the VAEP-controlled points regression
($p<10^{-4}$) and efficiency in the xG-difference regression ($p<10^{-4}$), with
VAEP non-significant in both. Repeated-team dependence is not further modelled.
On the spatial side, team assignment degrades for
low-contrast kits (white and striped kits, night matches): three windows were
excluded on this ground and one on clip quality, and per-match kit comparators
were set by inspection. Extra-time windows are excluded because regulation
kickoff offsets do not transfer. The SCI depends on pitch control computed over
imputed off-screen players; we did not filter windows on imputation sensitivity,
and a systematic sensitivity study across the full corpus is future work, as is
velocity-aware space \emph{value} (weighting controlled space by threat) and a
learned replacement for the operational thresholds.

\section{Reproducibility}
\label{sec:repro}

The event corpus is collected from public WhoScored match-event feeds for the
2026 FIFA World Cup; the spatial layer consumes the corresponding tournament
broadcasts. The event-side index, validation battery (including the VAEP
non-reducibility regression), and tournament aggregation run on CPU over this
event corpus and are released with code at
\url{https://github.com/nowayfootball/junk-possession}.
The spatial case-study pipeline consumes World Cup broadcast footage we cannot
redistribute; the pipeline code is described but the footage and per-clip
artifacts are not included. No GPU or cloud expenditure was used.

\bibliographystyle{plainnat}
\bibliography{refs}

\end{document}